\documentclass[runningheads]{llncs}

\usepackage[final,year=2026,ID=15]{eccv}

\usepackage{eccvabbrv}
\usepackage{graphicx}
\usepackage{float}
\usepackage{placeins}
\usepackage{booktabs}
\usepackage{amsmath,amssymb}
\usepackage{hyperref}

\newcommand{\colmap}{\mbox{COLMAP}}
\newcommand{\subcap}[2][2.0]{%
  \parbox[t][#1\baselineskip][t]{\linewidth}{\centering\small #2}}
\newcommand{\SEthree}{\mathrm{SE}(3)}
\newcommand{\SIMthree}{\mathrm{Sim}(3)}

\begin{document}

\title{When pose graphs are not enough: a failure analysis of
pose-graph-based camera estimation for Gaussian splatting}
\titlerunning{When pose graphs are not enough}

\author{Senthil Palanisamy \and Azwad Sabik \and Alan Dolhasz \and Mira Slavcheva}
\authorrunning{S. Palanisamy et al.}
\institute{Matterport}

\maketitle

\begin{abstract}
Exterior Gaussian-splat twins for real-estate listings are captured to a fixed
recipe: a consumer drone orbits a single property in several overlapping passes
at stepped altitudes, with occasional vertical ascents, so the imagery is dense,
sequential, GPS- and time-tagged, and revisits each facade from many angles.
This structure is tantalizing: it suggests camera poses could be recovered by a
\emph{dense pose-graph optimization} (PGO) whose only unknowns are the poses, far
fewer than the joint pose-and-point problem bundle adjustment (BA) solves.
PGO is mature in robotics, but almost always as \emph{post-processing}: it
redistributes the drift of an already-decent trajectory once loop closures tie it
together. Splatting captures push that regime to its limit, every frame closing
loops with dozens of others, raising a clean question: does pose-only
optimization reach BA accuracy? We built a system \emph{specialized to this
capture structure}, not a general-purpose SfM tool, to test it. It beat
\colmap{} on speed but never on quality; this paper is the candid post-mortem. The shortfall arises from several interacting causes,
hand-set weights, greedy merging and a noisy GPS ruler, together with two limits
inherent to the formulation: $\SEthree$ edges carry no scale degree of freedom,
so independently scaled sub-maps drift when merged, and a two-view edge stops
short of the coupling one 3D point imposes on every frame that sees it.
A viable design must add back $\SIMthree$ constraints, absolute GPS priors,
retained structure, and uncertainty-aware weighting.
\end{abstract}

\section{Introduction}
\label{sec:intro}
Gaussian splatting~\cite{kerbl3Dgaussians} has made high-quality novel-view
synthesis of real scenes routine, and with it the need for accurate camera poses
at scale. The default recipe passes images through a structure-from-motion (SfM)
system, almost always \colmap{}~\cite{colmap}, before training. For large
exterior scenes this SfM step is the bottleneck: robust incremental
reconstruction of thousands of drone frames can take hours.

The captures used for exterior splatting are, however, far from the
``images-in-the-wild'' regime SfM is built for: as above, a home is flown as
several dense overlapping orbits, every frame carries GPS, and consecutive frames
overlap strongly. It is natural to ask whether this structure lets us avoid the
expensive part of SfM. Bundle adjustment~\cite{triggs2000bundle} jointly refines
camera poses \emph{and} a large point cloud, its cost nominally dominated by the
points: each 3D point adds three parameters and there are far more points than
cameras. If we could constrain the cameras \emph{directly}, through a dense graph
of pairwise relative-pose measurements, the optimization would touch only the six
degrees of freedom per camera and be both smaller and trivially decomposable per
sequence.

\begin{figure}[t]
  \centering
  \includegraphics[width=\linewidth]{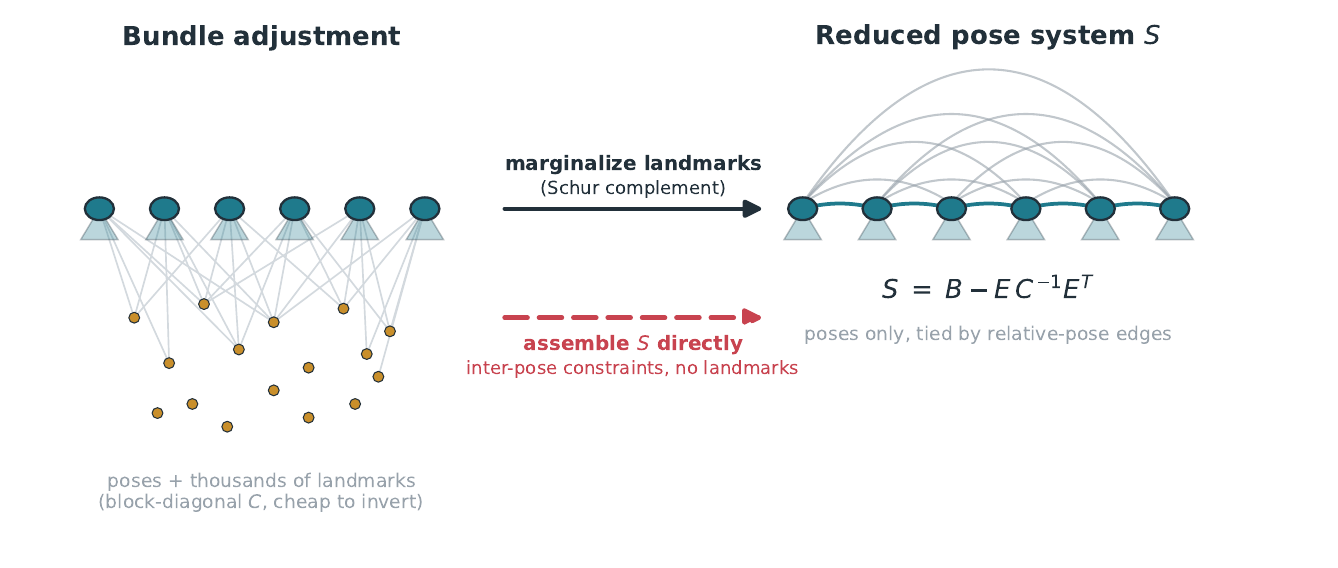}
  \caption{\textbf{The pose-only intuition.} Bundle adjustment (left) couples
  the camera poses to a large landmark cloud through observation edges. Because
  the landmark block $C$ is block-diagonal and cheap to invert, BA marginalizes
  the landmarks via the \emph{Schur complement} into a much smaller, pose-only
  reduced system $S = B - EC^{-1}E^{\top}$ (right), solves it for the poses, and
  only then back-substitutes for the points. Since BA's expensive core is
  already this pose-only system, a pose-only formulation tries to assemble $S$
  \emph{directly} from inter-pose relative-pose constraints (dashed), skipping
  the landmark detour. Both routes target the same $S$; this paper asks whether
  the shortcut recovers bundle-adjustment-grade poses.}
  \label{fig:schur}
\end{figure}

This intuition can be sharpened, and it is what makes a pose-only formulation
tempting rather than naive. Each Gauss--Newton step of bundle
adjustment~\cite{gao2021slam} solves a normal-equation system that, with the
camera update $\Delta\mathbf{x}_c$ ordered above the landmark update
$\Delta\mathbf{x}_p$, takes the block form
\begin{equation}
\begin{bmatrix} B & E \\ E^{\top} & C \end{bmatrix}
\begin{bmatrix} \Delta\mathbf{x}_c \\ \Delta\mathbf{x}_p \end{bmatrix}
=
\begin{bmatrix} \mathbf{v} \\ \mathbf{w} \end{bmatrix},
\label{eq:ba}
\end{equation}
with a camera block $B$, a landmark block $C$, and their coupling $E$. Because
$C$ carries one $3\times3$ diagonal block per 3D point it is block-diagonal and
cheap to invert, so BA does not in fact pay for the landmarks directly: it
eliminates $\Delta\mathbf{x}_p$ by the Schur complement, left-multiplying
\cref{eq:ba} by $\left[\begin{smallmatrix} I & -EC^{-1}\\ 0 &
I\end{smallmatrix}\right]$ to reach the much smaller, pose-only reduced system
\begin{equation}
\underbrace{\big(B - E C^{-1} E^{\top}\big)}_{\textstyle S}\,\Delta\mathbf{x}_c
= \mathbf{v} - E C^{-1}\mathbf{w},
\label{eq:schur}
\end{equation}
and only then back-substitutes $\Delta\mathbf{x}_p = C^{-1}(\mathbf{w} -
E^{\top}\Delta\mathbf{x}_c)$. Forming the reduction and back-substituting are
redone at every iteration, so the dominant work is the repeated solve of $S$, not
the landmarks. If BA's expensive core is already a pose-only system, the
attractive idea is to build $S$ \emph{directly}, as a pose graph of relative-pose
constraints, skipping the landmark bookkeeping and marginalization altogether and
arriving at the reduced system for free (\cref{fig:schur}). The core question is
whether the $S$ that pose-graph
optimization assembles directly approaches the $S$ that BA produces by
marginalization, in the limit of increasingly many inter-frame constraints.

There is a further, more principled reason to study this here. Pose-graph
optimization is a mature and well-understood tool~\cite{grisetti2010tutorial}, but in robotics and SLAM~\cite{cadena2016past} it is
almost always a \emph{back-end refinement}: a front end (visual odometry or SLAM)
first produces poses of reasonable accuracy, and PGO then uses loop-closure edges
to redistribute the accumulated drift into a globally consistent trajectory. Its
power grows with the number and diversity of those loop closures.
Splatting captures sit at an extreme of that regime: because the imagery is so
dense, essentially every frame closes loops with dozens of others, so the pose
graph is far richer than the sparse loop-closure graphs PGO is usually handed.
This invites a genuine research question, \emph{in the limit of an ever-denser
loop-closure graph, does pose-only optimization converge to bundle-adjustment
accuracy?} If it did, the dense structure of splatting captures would be exactly
the setting in which to trade BA for PGO. We treat answering this, empirically
and then structurally, as the study's core purpose.

This is the hypothesis we set out to validate: \emph{that dense pose-graph
optimization, used not as a post-hoc SLAM refinement but as the primary estimation
objective, can recover poses of BA-grade quality \emph{on these structured,
GPS-tagged drone captures}, faster and in a cloud-parallel form}. We built the system and ran it on self-captured drone
scenes and a public benchmark. The strong form of the hypothesis proved false.
This paper explains why, using the system's own design and numbers, and is
careful about which parts of the explanation generalize: two of the causes are
properties of the pose-only formulation that no parameter setting escapes, while
the rest are choices this pipeline made and a better one could revisit. We keep
the two apart throughout (\cref{sec:why}).

We recall the
related landscape (\cref{sec:related}), the pipeline (\cref{sec:system}) and its
outcomes (\cref{sec:outcomes}), then give the methodological failure analysis
that is our contribution (\cref{sec:why}) and the design lessons it implies
(\cref{sec:lessons}).

\section{Related work}
\label{sec:related}
\colmap{}~\cite{colmap} is the de facto \emph{general-purpose} standard for pose
estimation in vision and the default front end for splatting. We do not aim to
replace it there: we target only the narrow, structured regime of exterior
splatting captures, using \colmap{} as the reference a specialized pose-only
design must match. GLOMAP~\cite{pan2024glomap} showed that
global SfM can approach incremental accuracy at lower cost, yet \colmap{} remains
the method of choice when robust registration is required.
Theia~\cite{theia-manual} offers both incremental and global SfM but is chiefly a
researcher's toolkit. Learned, feed-forward estimators such as
VGGT~\cite{wang2025vggt} and Depth-Anything-3~\cite{depthanything3} predict poses,
depth and intrinsics in a single pass and are advancing quickly. They make
excellent one-shot pose initializers, but as we confirm below they have not yet
reached the pose accuracy of a robust system that nonlinearly refines poses.

Our matching front end follows hierarchical
localization~\cite{sarlin2019coarse}, pairing SuperPoint
features~\cite{super} with LightGlue~\cite{lindenberger2023lightglue} and
NetVLAD~\cite{Arandjelovicetal2016} retrieval indexed with
FAISS~\cite{douze2024faiss}, a modern replacement for
bag-of-words retrieval~\cite{Liu2016BoVW}. Distributing SfM across machines dates
back to Building Rome in a Day~\cite{agarwal2009building}. Our design differs in
demoting bundle adjustment~\cite{triggs2000bundle} from the global objective to a
per-chunk step and making \emph{pose-graph optimization} carry the estimation,
solved with g2o~\cite{kuemmerle2011g2o} (GTSAM~\cite{gtsam} is the common
alternative), and in exploiting GPS and sequential capture. It is the limits of
that partial substitution, and of the pure-pose formulation it approximates,
that we analyze.

\section{The system we built}
\label{sec:system}
\begin{figure}[ht]
\centering
\includegraphics[width=0.62\linewidth]{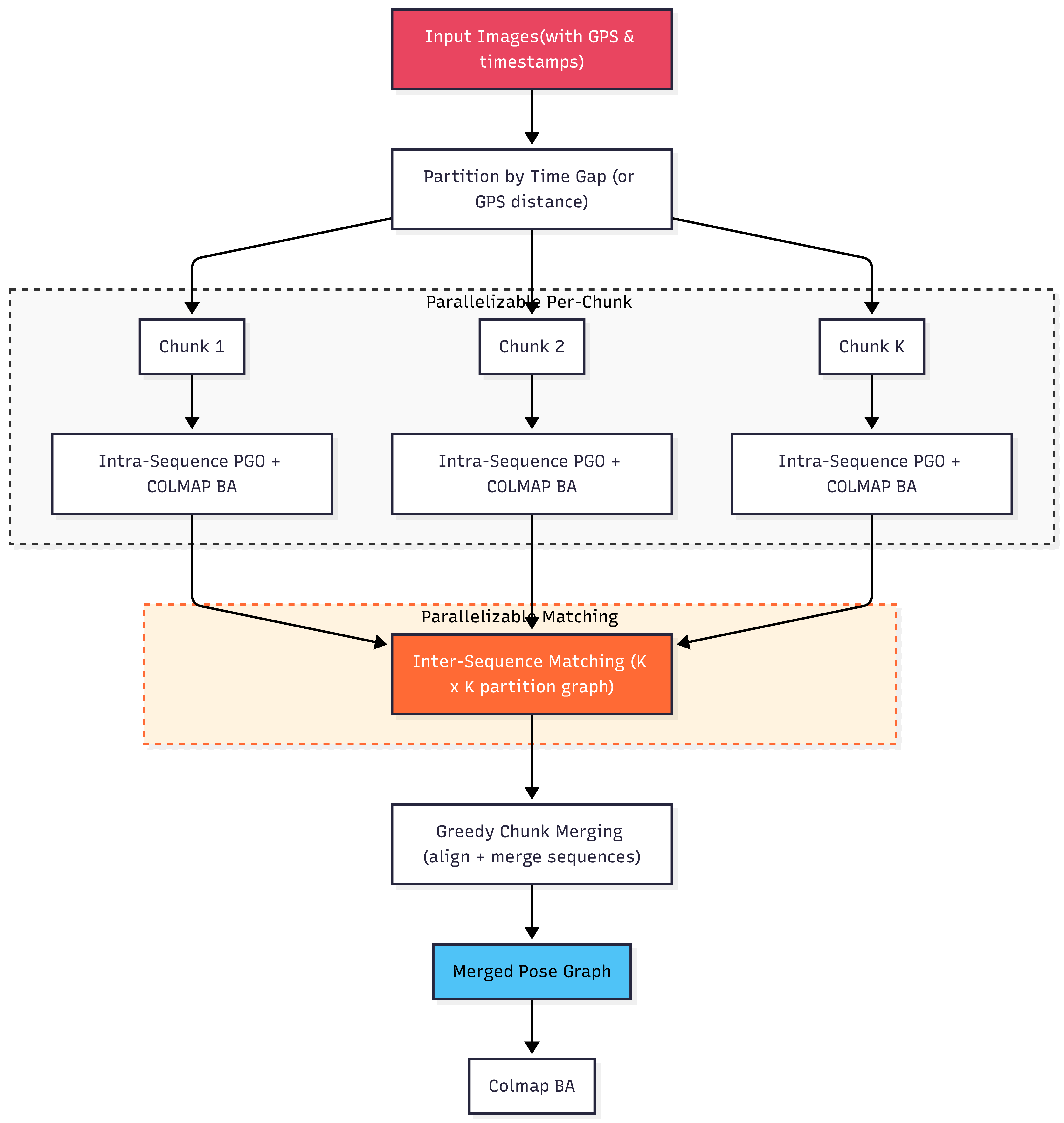}
\caption{System overview. The image stream is split into per-sequence
\emph{chunks} that are optimized independently as dense $\SEthree$ pose graphs,
then merged pairwise into a single graph. Every chunk-level solve is independent,
which is what the design intended to parallelize across the cloud.}
\label{fig:overview}
\end{figure}

We summarize the pipeline only enough to make the analysis precise. It is not a
general-purpose SfM system but one built expressly around the structure of these
captures, dense sequential overlap, per-frame GPS, and multi-orbit coverage; a
three-stage divide--compute--merge design (\cref{fig:overview}).

\paragraph{Chunks.}
The image stream is split by timestamp into disjoint \emph{chunks}
(``sequences''), each a continuous imaging run in which every frame overlaps
another frame in the same chunk. Chunks larger than $200$ frames are subdivided.
Chunks are the unit of parallelism.

\paragraph{Intra-sequence PGO.}
Within a chunk, images are indexed with NetVLAD~\cite{Arandjelovicetal2016} in a
FAISS~\cite{douze2024faiss} store; each frame retrieves visual neighbors plus
GPS-proximate geometric neighbors. Retrieved pairs are matched with
SuperPoint~\cite{super}$+$LightGlue~\cite{lindenberger2023lightglue}, verified
against degeneracy (minimum feature motion, essential-vs-homography inlier test,
a $3^\circ$ median-triangulation-angle floor), and reduced to a relative rotation
and an up-to-scale translation from the essential matrix~\cite{nister2004five}. That translation is
scaled by the GPS distance between the two frames. Each verified pair becomes an
edge in an $\SEthree$ pose graph solved with g2o~\cite{kuemmerle2011g2o}; the
objective is the standard robust sum of relative-pose residuals,
\begin{equation}
E=\!\!\sum_{(i,j)\in\mathcal{E}}\!\!
\rho\!\left(\big[\log(T_j^{-1}T_{ji}T_i)\big]^{\!\vee\top}
\,\Omega_{ji}\,
\big[\log(T_j^{-1}T_{ji}T_i)\big]^{\!\vee}\right),
\label{eq:pgo}
\end{equation}
with $\Omega_{ji}$ a hand-set information matrix (high on rotation, low on
translation) and $\rho$ a Huber kernel. Each chunk yields a dense graph in which
a frame is tied to about $100$ others.

\paragraph{Retained bundle adjustment.}
The pipeline does not end there, and this is the design's most consequential
compromise. Each chunk's matches are imported into a \colmap{} database and a
bundle-adjustment pass refines poses, intrinsics, and points. The points that
pass triangulates are then reused: they are what the $3$D--$2$D constraints
below are built from. BA is thus removed only \emph{globally}. It is not run
over the full capture, which is where the cost argument of \cref{sec:intro}
bites, but it survives inside every chunk, and structure survives with it. The
system is accordingly BA-reduced, not BA-free, and we return
in \cref{sec:why-structure} to why the pure-pose design could not shed this
last dependency.

\paragraph{Inter-sequence constraints and merge.}
Chunks are still disconnected. We form a $K\times K$ table whose $(i,j)$ entry
holds cross-chunk constraints, $2$D--$2$D essential (GPS-scaled) and
$3$D--$2$D perspective-$n$-point (PnP)~\cite{lepetit2009epnp} edges, the latter
resolved against the per-chunk BA points; $3$D--$3$D point-cloud alignment was
tried and abandoned as unstable. The final stage greedily merges the pair of
chunks with the most constraints: a robust $\SEthree$ average of the pairwise
relative transforms (iterated $3\sigma$ rejection) pre-aligns one chunk to the
other, g2o re-optimizes with the cross-chunk edges enabled, and the two chunks
collapse into one. This repeats until a single graph remains. Information
matrices encode trust: PnP edges $7\times$ essential edges, intra-sequence edges
up-weighted $10\times$ before the merge.

\paragraph{Intended parallelism.}
Every intra-sequence solve is independent and each $K\times K$ entry is
independent, so the front end is designed to run across a cloud cluster.

\section{Outcomes}
\label{sec:outcomes}
\begin{figure}[tbp]
\centering
\begin{minipage}{0.30\linewidth}\centering
\includegraphics[width=\linewidth]{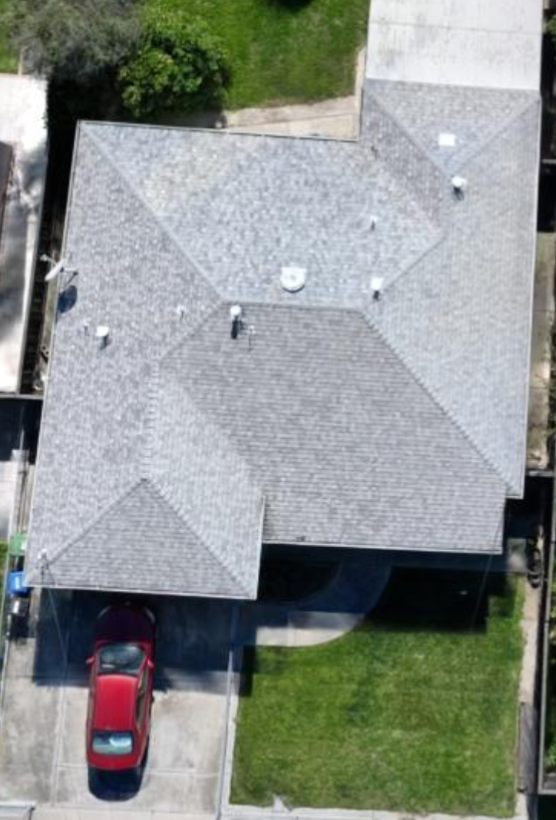}\\[2pt]
\subcap{(a) Gaussian-splat from \colmap{} poses}
\end{minipage}\hspace{0.04\linewidth}
\begin{minipage}{0.30\linewidth}\centering
\includegraphics[width=\linewidth]{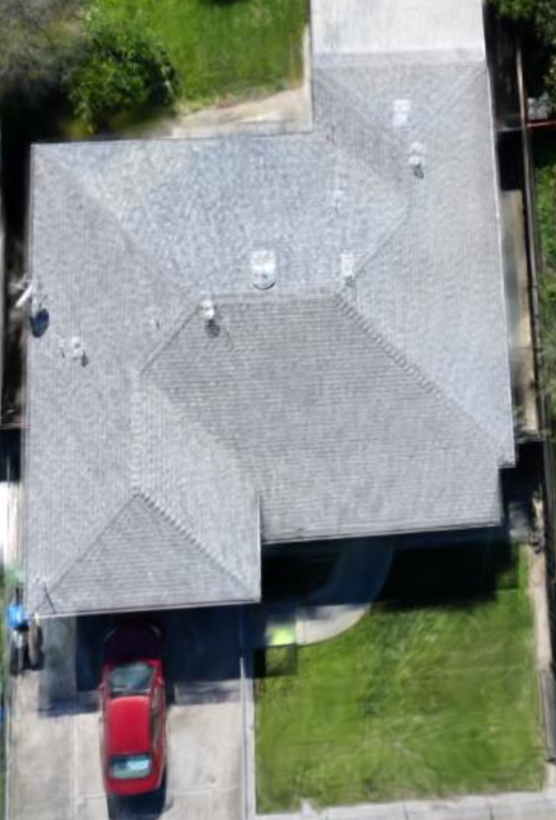}\\[2pt]
\subcap{(b) Gaussian-splat from PGO poses}
\end{minipage}\\[6pt]
\begin{minipage}{\linewidth}\centering
\includegraphics[width=\linewidth]{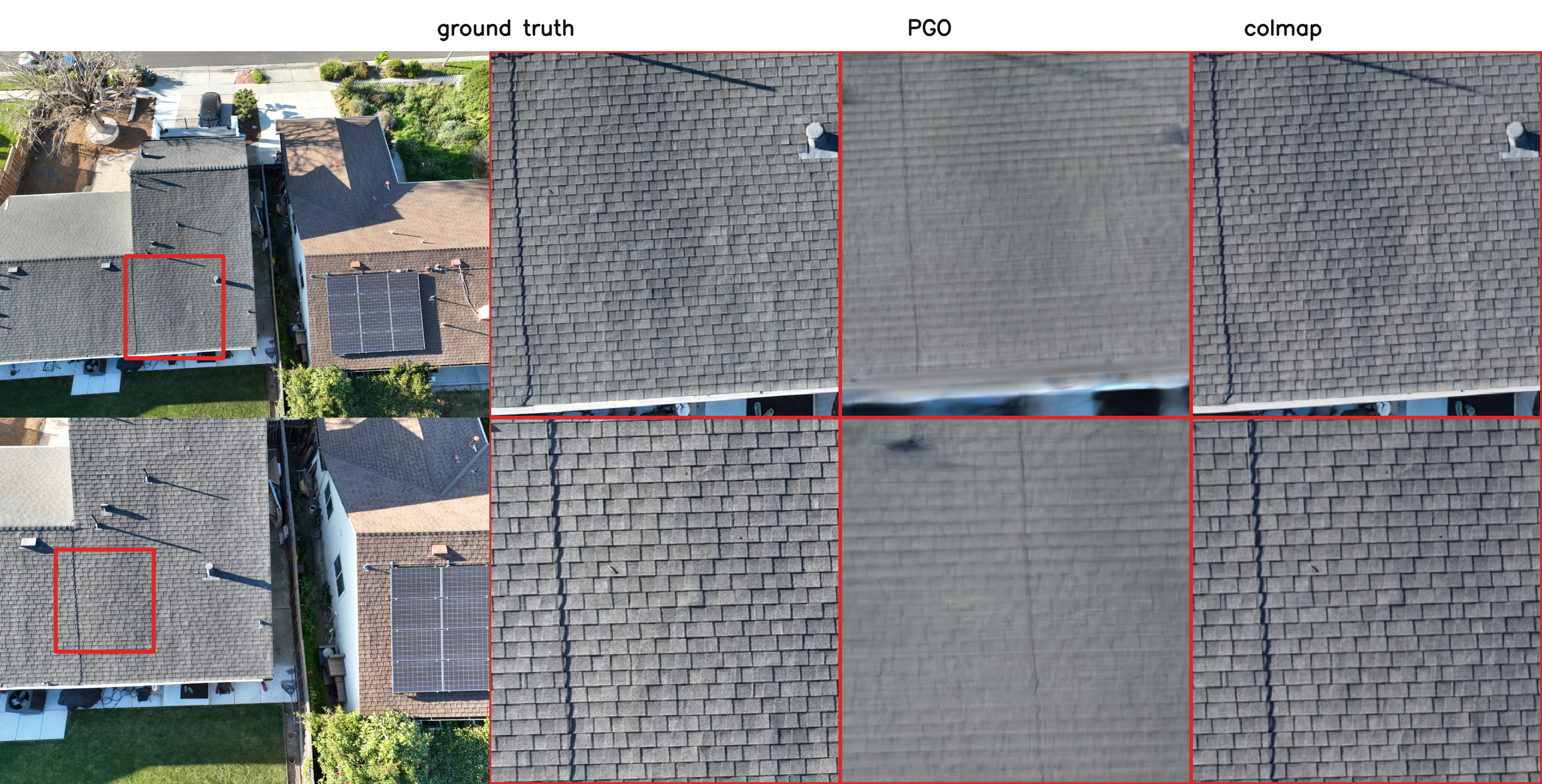}\\[2pt]
{\small (c) house~2: zoomed roof crops (red box) of the PGO and \colmap{}
Gaussian-splats against ground truth}
\end{minipage}
\caption{Splat quality. On house~1 the pose graph places cameras close to
\colmap{} ($0.5^\circ$, $2$\,m), yet the PGO reconstruction (b) is visibly
softer than \colmap{}'s (a), the seam blur left by scale drift across chunk
merges (\cref{sec:why-scale}) that a shared point cloud would have prevented.
Row (c) shows the same on a second scene (house~2): in the zoomed roof crops
the PGO splat (center) smears the shingle texture, while the \colmap{} splat
(right) resolves individual shingles, matching ground truth (left).}
\label{fig:splatcompare}
\end{figure}

\begin{figure}[tbp]
  \centering
  \begin{minipage}[b]{0.42\linewidth}\centering
    \includegraphics[width=\linewidth]{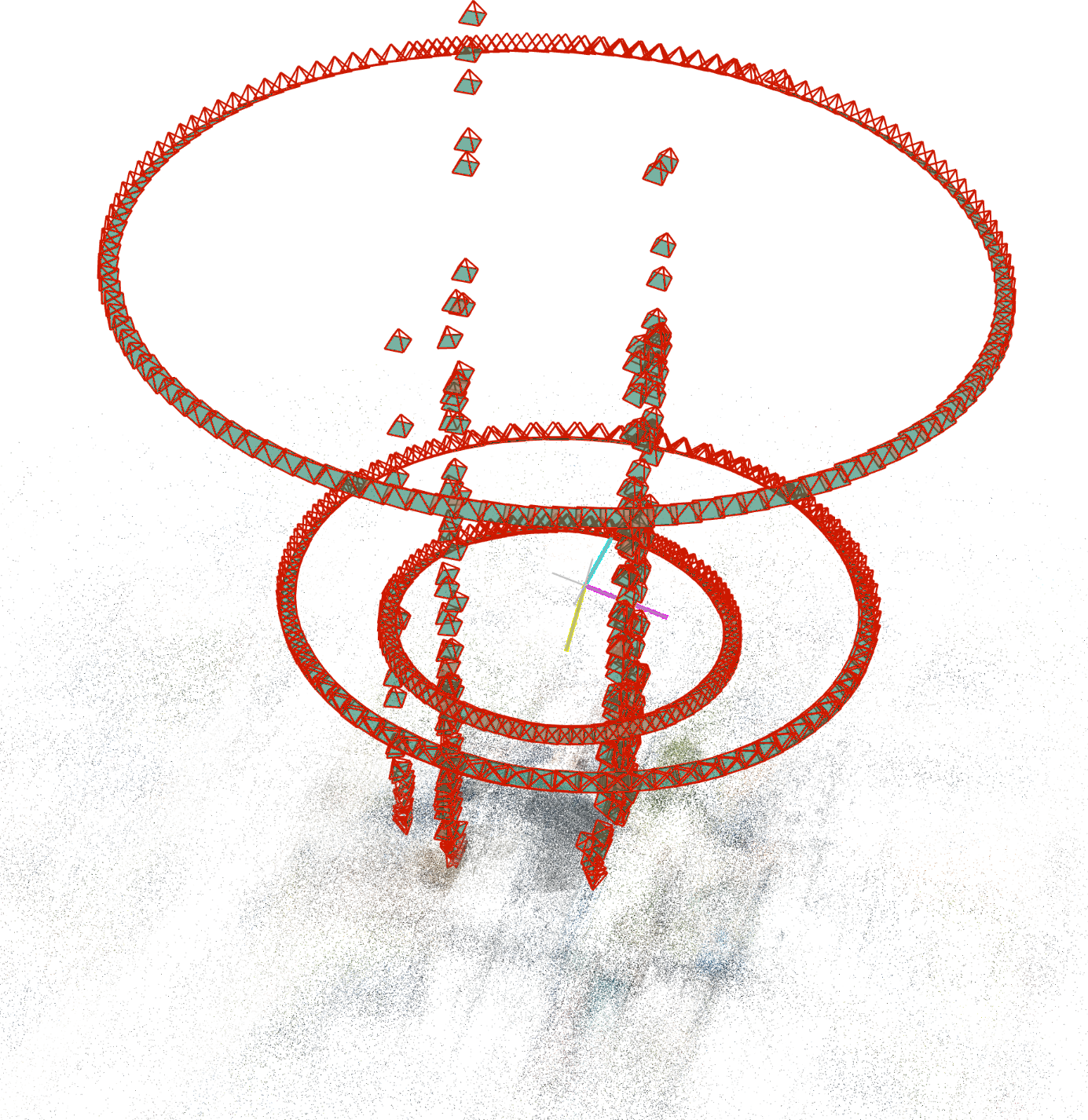}\\[3pt]
    \subcap[1.2]{(a) estimated camera poses}
  \end{minipage}\hfill
  \begin{minipage}[b]{0.42\linewidth}\centering
    \includegraphics[width=\linewidth]{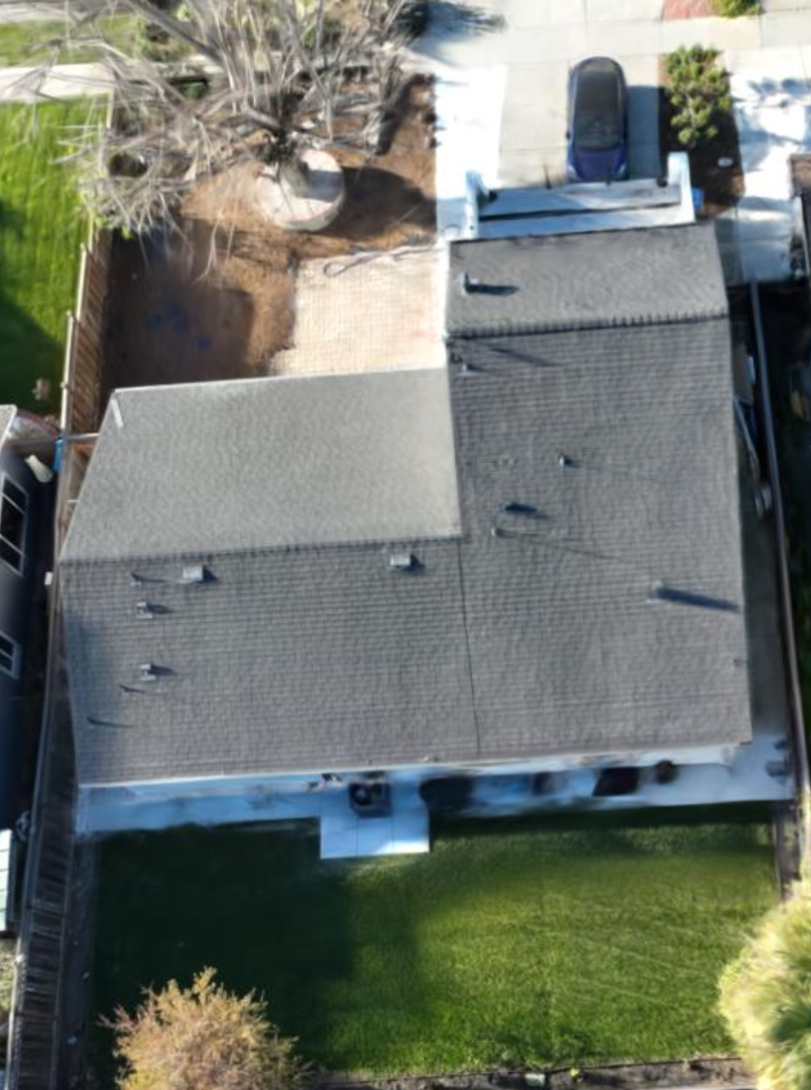}\\[3pt]
    \subcap[1.2]{(b) Gaussian-splat reconstruction}
  \end{minipage}
  \caption{Qualitative output of the pose-only pipeline on a self-captured house:
  (a) the camera poses it estimates and (b) the resulting Gaussian-splat
  reconstruction. The poses are close to \colmap{} and the splat is recognizable;
  the failures analyzed below are quality and consistency gaps, not gross
  breakage.}
  \label{fig:qualitative}
\end{figure}

We evaluated the pipeline internally on $40$ proprietary drone scenes. Across the
set it neither beats \colmap{} nor yields metrics useful on their own, so we do
not tabulate aggregate numbers that would only restate a uniform shortfall. Instead we elucidate two
representative self-captured scenes (houses of $824$ and $741$ frames) in detail:
precisely because they neither beat \colmap{} nor behave anomalously, they make a
clean test bed for reasoning about \emph{why} a pose-only, structure-free
formulation falls short. We compare against a tuned \colmap{} baseline (GPU
extraction and matching, vocab-tree matcher, hierarchical mapper), and add the
public GauU-Scene~V2~\cite{xiong2024gauuscene} drone dataset with lidar ground
truth.
\cref{tab:houses,tab:gauu} collect the numbers and
\cref{fig:splatcompare} shows a representative splat pair. The poses agree with
\colmap{} to $0.5$--$0.6^\circ$ and to within about $2$\,m, in a space where the
trajectories themselves span more than $100$\,m, and
the splats are recognizable (\cref{fig:qualitative}). Of the two axes the system
was built to win, speed and splat quality, the outcome is split. It has
real potential to beat \colmap{} on speed: the parallel time
beats \colmap{} by a wide margin. But we had no incentive to pursue that
implementation with full vigor, because the splat quality was never comparable to
what \colmap{} poses produce. On the external dataset one scene failed outright and the rest contained large
outliers, though we do report the median here. \cref{sec:why} explains
why each of these follows from the method's design.

\begin{table}[t]
\centering
\caption{Self-captured houses. Pose deviation is relative to \colmap{}. ``Wall''
is measured single-machine runtime; ``Par'' is the parallel time, the runtime
the (independent) front end achieves when its steps run in parallel.}
\label{tab:houses}
\setlength{\tabcolsep}{4pt}
\small
\begin{tabular}{lc cc ccc cc}
\toprule
 & & \multicolumn{2}{c}{dev.\ vs \colmap} & \multicolumn{3}{c}{runtime [min]}
   & \multicolumn{2}{c}{splat PSNR\,$\uparrow$\,/\,FID\,$\downarrow$} \\
\cmidrule(lr){3-4}\cmidrule(lr){5-7}\cmidrule(lr){8-9}
Scene & \#img & Rot$^\circ$ & Trans\,m & Wall & Par & \colmap & ours & \colmap \\
\midrule
house\,1 & 824 & 0.5 & 2.0 & 97 & 37 & \textbf{70} & $16.71/0.93$ & $\mathbf{18.38/0.33}$ \\
house\,2 & 741 & 0.6 & 2.1 & 95 & 33 & \textbf{58} & $14.95/3.24$ & $\mathbf{17.60/0.61}$ \\
\bottomrule
\end{tabular}
\end{table}

\begin{table}[t]
\centering
\caption{GauU-Scene~V2~\cite{xiong2024gauuscene}: a reconstructed trajectory
(left) and per-scene deviation from the lidar ground truth (right; medians,
means skewed by large outliers; ``Failed''\,=\,no usable reconstruction).}
\label{tab:gauu}
\begin{minipage}[c]{0.44\linewidth}\centering
  \includegraphics[width=\linewidth]{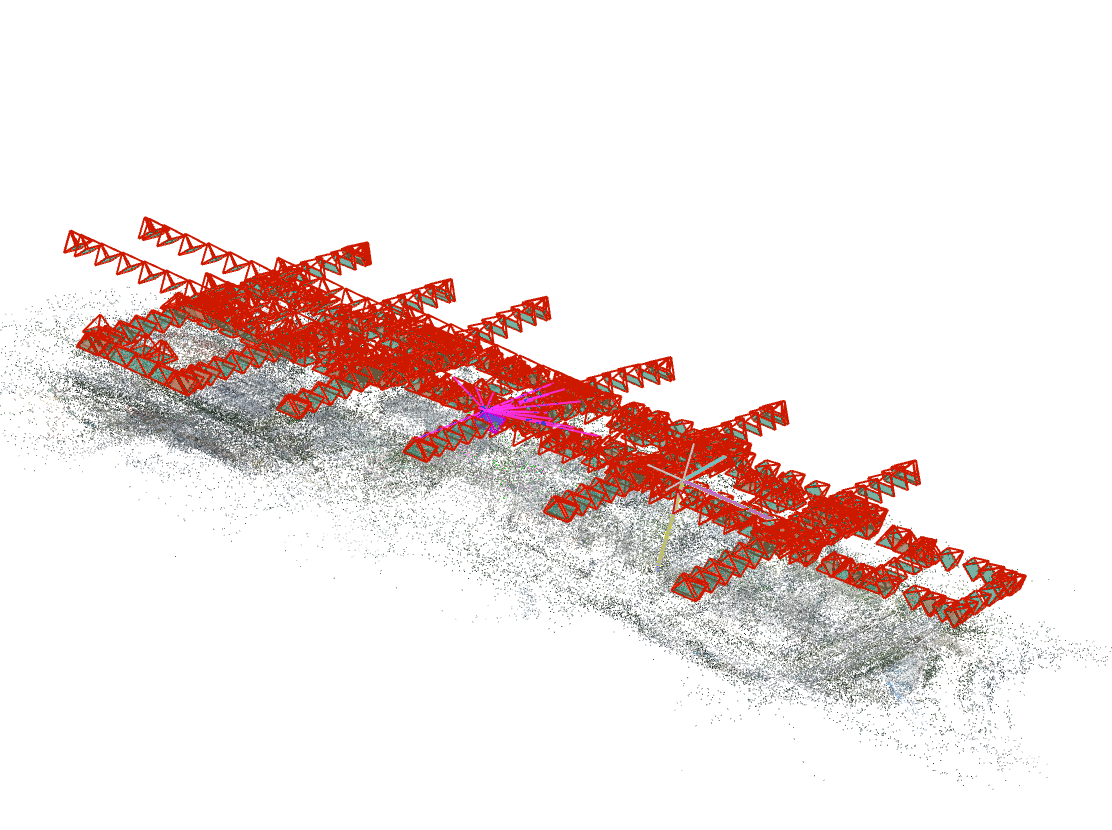}
\end{minipage}\hfill
\begin{minipage}[c]{0.52\linewidth}\centering
  \setlength{\tabcolsep}{6pt}
  \begin{tabular}{lcc}
  \toprule
  Scene & Rot$^\circ\downarrow$ & Trans\,m$\downarrow$\\
  \midrule
  CUHK\_UPPER & 0.8 & 4.2\\
  HAV & \multicolumn{2}{c}{\textbf{Failed}}\\
  LFLS & 1.13 & 6.3\\
  \bottomrule
  \end{tabular}
\end{minipage}
\end{table}

\section{Why the method fails}
\label{sec:why}
We give four causes, each supported by the system's own design and measurements.
They are not of one kind, and separating them matters for how much the reader
should conclude.

Two are \emph{representational}: they follow from what a pose-only formulation
can express, hold for any parameter setting, and are argued below from the
structure of the estimator rather than from our numbers. $\SEthree$ edges have no
scale degree of freedom (\cref{sec:why-scale}), and a two-view edge cannot encode
the simultaneous coupling that one point observed in many views imposes
(\cref{sec:why-structure}). No information matrix, merge order, or
initialization repairs either; they bound what the formulation can do.

The other two are \emph{implementation choices} of this pipeline: sequence-wise
ordering and merging (\cref{sec:why-order}) and reliance on consumer GPS as the
metric ruler (\cref{sec:why-gps}), together with the hand-set information
matrices of \cref{sec:why-trust}. Better choices exist for each, and we say so
as we go.

The honest limit of our evidence is the boundary between these two kinds. Since
we ablate none of the four, the analysis locates where a ceiling comes from
without showing that our measurements sit at it rather than at a lower floor set
by the implementation; a better-engineered pose-only pipeline should beat the
numbers in \cref{sec:outcomes}, perhaps substantially. What it cannot do is
escape the two representational limits, and \cref{sec:lessons} is written for
whoever tries.

\subsection{Runtime is not where the method fails}
The founding motivation was a runtime argument, and on its own terms it is
sound. The cost splits into two parts that scale very differently. The matching
front end, per frame, NetVLAD retrieval, SuperPoint$+$LightGlue against about
$100$ neighbors, and two-view verification of each pair, is the dominant
\emph{absolute} cost, but it is \emph{embarrassingly parallel}: every pair is
matched and verified independently, so it shards trivially across a cloud fleet
and imposes no fundamental serial bottleneck. What does \emph{not} parallelize is
the \emph{solver}, the global optimization that ties the frames into one
consistent model, and that is the fundamental bottleneck a distributable
pipeline must attack. Reducing it is exactly what a pose-only formulation sets
out to do: shrink the per-step system to the six-DoF cameras (\cref{eq:schur})
and solve each chunk independently, making the one part that cannot be
parallelized as small and as decomposable as possible. In that sense the design
targets the \emph{right} cost. The single-machine (sequential) runtime in \cref{tab:houses}
does not beat \colmap{}, but the parallel time does. Speed, then, is not where this
method fails; its failures, examined next, are geometric.

\subsection{$\SEthree$ edges cannot represent scale, so merged sub-maps drift}
\label{sec:why-scale}
This is the deepest failure. Each intra-sequence graph is metric only up to the
GPS scaling applied to its essential-matrix translations, and different chunks
acquire slightly different effective scales because their GPS baselines and
triangulation geometry differ. The merge then aligns chunks with an $\SEthree$
transform, and \cref{eq:pgo} lives entirely in $\SEthree$, which has \emph{no
scale degree of freedom}. When two chunks disagree on scale by even a few
percent, an $\SEthree$ alignment cannot reconcile them: it trades the error into
translation and leaves a residual the optimizer cannot drive to zero. The visible
symptom is blur and ghosting at the seams of the merged splat
(\cref{fig:splatcompare}b), and it contributes to the $1.7$--$2.7$\,dB PSNR
deficit in \cref{tab:houses}, alongside the GPS noise and weighting choices
discussed below, which our measurements cannot separate from it. The argument
does not need the attribution: the limitation is visible in the state space
itself, before any measurement. Bundle adjustment never encounters this, because 3D
points shared across overlapping views implicitly tie their scales together. A
pose graph in $\SEthree$ discards that coupling and has no other mechanism to
recover it, so scale drift across merges is not a bug to be tuned away; it is
unrepresentable in the chosen state space.

\begin{figure}[tbp]
\centering
\includegraphics[width=\linewidth]{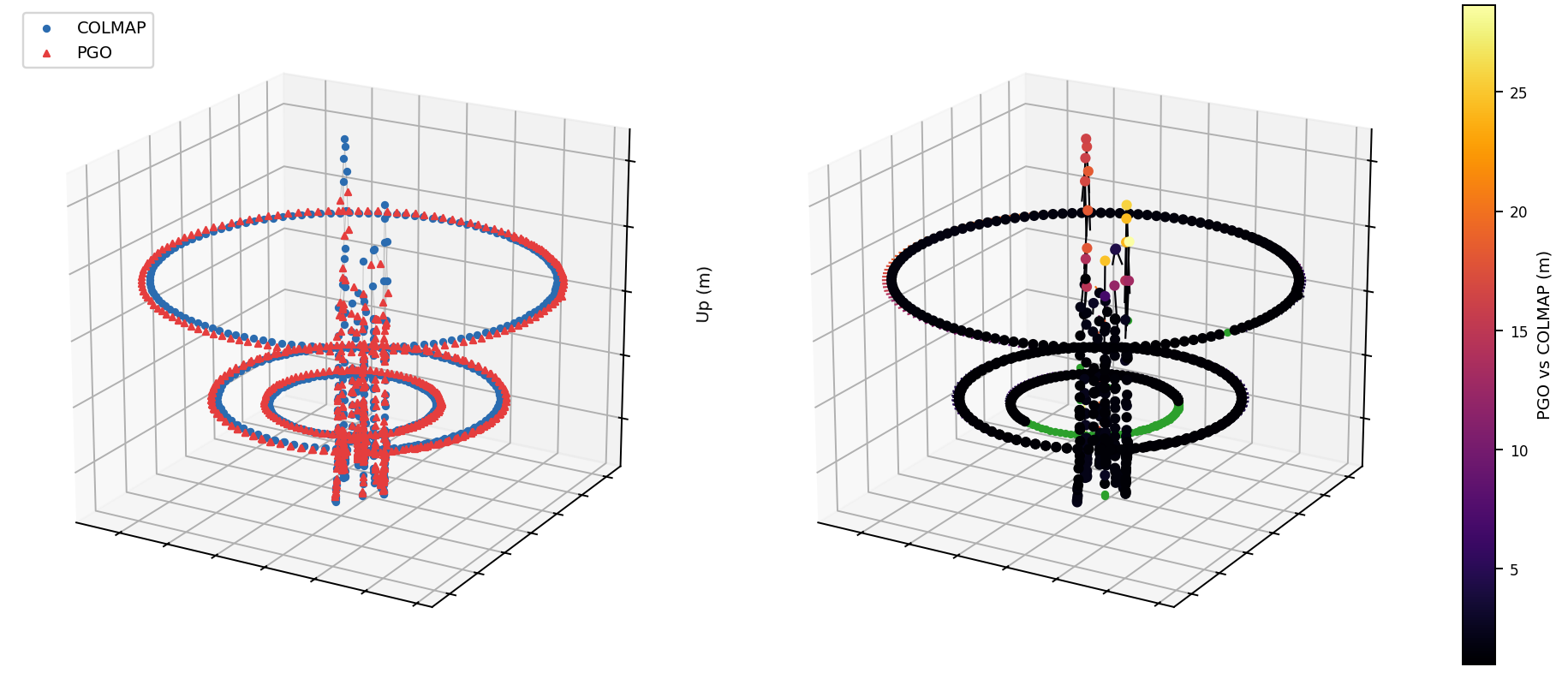}
\caption{Camera centers, PGO versus \colmap{}, both aligned to the GPS metric
frame (house~2, $741$ images, a $148$\,m scene with $104$\,m of vertical
ascent). Left: the two sets of centers overlaid. Right:
every frame that disagrees by more than $1$\,m is highlighted in color, the
rest green. The disagreement is pervasive, $90\%$ of frames exceed $1$\,m
(median $1.2$\,m), and worst on the vertical-ascent frames, up to $29$\,m. The
two regimes fail differently: the near-linear ascent trajectories suffer nadir
degeneracies, while the rest of the scene inherits scaling error from noisy
GPS, either way enough to soften the PGO splat relative to \colmap{}.}
\label{fig:posediff}
\end{figure}

Restoring the missing degree of freedom does not rescue the result, and we tried:
lifting each merge to $\SIMthree$, which \emph{does} carry a scale term. Such a
merge must estimate one similarity $(s, R, t)$ per sequence boundary from the
inter-sequence constraints alone, and with few, wide-baseline shared observations
that estimate starts far from its optimum. Compounding this, the GPS-derived
constraints \emph{conflict}: each pulls the boundary toward a different,
meter-scale, anisotropically noisy anchor, so the solve settles at a compromise
satisfying none of them, metrically plausible but still blurring the seam. The
missing scale degree of freedom is therefore necessary but not sufficient.

Under a sequential merge this conflict is structural rather than an artifact of
noise levels, and seeing why shows exactly where $\SIMthree$ alone relocates the
problem instead of removing it. Scaling an entire sequence
together with all of its pose constraints does keep that sequence internally
consistent, so once we scale sequence~$2$ to agree with sequence~$1$ the pair
$(1,2)$ is mutually consistent, and every inter-sequence constraint incident on
$2$ is scaled to match. An astute reader might expect this rescaling to be
absorbed automatically as the next sequence is merged. It is not. When a third
sequence is merged against the first two, the inter-pose constraints
$3\!\to\!1$ and $3\!\to\!2$ both carry the raw GPS scale, because that is how
sequence~$3$ was reconstructed. But sequence~$2$ has already been rescaled to
match~$1$, so $3\!\to\!1$ is roughly consistent up to ordinary scale drift while
$3\!\to\!2$ disagrees by exactly the correction we applied to~$2$. The optimizer
is handed two contradictory statements about the same scale, and no single
scalar satisfies both. The similarity that resolved $(1,2)$ thus reappears as a
fresh conflict at~$3$: $\SIMthree$ shifts the scale mismatch along the merge
order rather than removing it.

\paragraph{Multi-way synchronization was not attempted.}
The contradiction above arises only because the merges happen \emph{one after
another}: each merge aligns the incoming chunk to the partial map so far, so by
chunk~$3$, chunk~$2$ has already been corrected by its merge with~$1$ while
chunk~$3$ still carries raw GPS scale. That mismatch, not $\SIMthree$ itself, is
what no single scalar can satisfy.

The alternative is to estimate all $K$ chunk similarities \emph{at once} from the
full graph of inter-chunk constraints, so no chunk is corrected before the others
are accounted for. We did not try it. Such multi-way $\SIMthree$ synchronization,
in the spirit of motion and rotation
averaging~\cite{govindu2001combining,hartley2013rotation}
and similarity synchronization~\cite{arrigoni2020synchronization}, spreads
inconsistent pairwise scales in a least-squares sense over all chunks.

The scale component in particular is a \emph{linear} problem, which is what makes
this attractive. Give chunk~$k$ an unknown log-scale $\sigma_k=\log s_k$; each
cross-chunk constraint between $i$ and $j$ measures a ratio $s_{ij}$ and so
contributes one equation $\sigma_j-\sigma_i=\log s_{ij}$. Over the populated
entries of the $K\times K$ table these stack into an overdetermined linear system
in $K$ unknowns whose normal equations are a weighted graph Laplacian, determined
up to a single global gauge that GPS, or fixing one chunk, supplies. Rotations
follow by rotation averaging and translations are linear once scales and
rotations are known, the standard $\SIMthree$ synchronization
recipe~\cite{arrigoni2020synchronization}. Crucially, every chunk's scale is then
set by \emph{all} of its constraints simultaneously, so a disagreement between
$3\!\to\!1$ and $3\!\to\!2$ is distributed as residual rather than resolved in
favour of whichever merge happened first. With $K$ in the tens the solve is
negligible beside the matching front end, so it costs nothing the speed argument
cares about. We expect this to reduce the seam error materially and regard it as
the most promising untried repair, though it redistributes inconsistency rather
than supplying the absolute scale the capture lacks.

\subsection{Sequence-by-sequence ordering leaves sub-maps ill-conditioned before
they merge}
\label{sec:why-order}
This cause and the two that follow are implementation choices, each of which
could have been made differently and none of which we ablated. They are worth
recording because the divide-and-merge decomposition that makes a pose-only
design attractive to parallelize is exactly what makes them tempting.

The pipeline orders images by capture sequence, solves each sequence
independently, and only then merges, and the ordering itself is a failure
mode, for two compounding reasons. First, some sequences are \emph{geometrically
degenerate in isolation}: a nadir ascent, where the drone rises vertically over a
fixed point, supplies almost no translational parallax, so the sub-map is
\emph{fundamentally unconstrained} up to the missing baseline. Solved alone, it
has no way to recover the structure a neighbouring oblique sequence would have
supplied. Second, to absorb the meter-scale GPS noise each sequence settles into
its own \emph{local optimum} that best trades its observations against its noisy
anchors. Because every sub-map made that trade independently, the local optima
disagree; when the sequences merge, the cross-sequence constraints are in
\emph{large mutual conflict}, far outside the basin the merge can reconcile, and
the global solve settles at a sub-optimal compromise.

The lesson is about ordering, not merging: an image ordering that proceeds
strictly sequence-by-sequence guarantees these isolated, separately-converged
pieces, and no merge strategy applied afterwards can undo a sub-map that has
already converged to the wrong basin. An ideal ordering would never take one
sequence to convergence in isolation. Interleaving images \emph{across} sequences
from the start, so that cross-sequence constraints shape each pose before it
commits to a local optimum, keeps the whole capture in a single jointly
constrained basin rather than assembling it from pre-settled, mutually
inconsistent parts, and it costs nothing in parallelism: the chunking that makes
the front end shardable is independent of the order in which poses are allowed to
influence one another. We did not build that ordering, so we cannot say how much
of the merge conflict it would remove; it is the ablation we would run first.

\subsection{Removing shared structure removes multi-view consistency}
\label{sec:why-structure}
The premise was that pose-only optimization could reach BA-grade quality. But the
thing that makes BA's poses globally consistent is precisely the structure the
method discards: a 3D point observed in many frames forces those frames to agree,
and BA averages their independent measurement errors into one consistent
estimate. A pose graph of pairwise relative transforms has no such shared latent
variable; it is only as good as its edges, and edge errors accumulate rather than
cancel. Two facts in the system confirm that this coupling could not simply be
dropped. First, the pipeline still runs a per-chunk \colmap{} bundle adjustment,
and the $3$D--$2$D PnP constraints that stitch chunks are derived from the points
that BA produced, so structure was reintroduced through the back door. This cuts
two ways, and we state both. It is the clearest evidence that the coupling is
load-bearing: the design reached for structure exactly where pose-only
constraints proved too weak to stitch sub-maps, and no version of the pipeline
ran without it. But it also means we never put the pure-pose premise fully on
trial, and our negative result is correspondingly bounded. What we have shown to
fail is pose-only estimation \emph{at the global scale}, with BA still
supporting each chunk from below; a formulation that removed BA entirely would
inherit every weakness catalogued here and lose that support besides, so we
expect it to fare worse, but we did not measure it. Second, the one path that tried to
use structure-to-structure information directly, the $3$D--$3$D point-cloud
alignment edge, was abandoned as unstable. That instability is no accident:
projectively triangulated points carry an uncertainty that grows with observation
distance and shrinks with disparity, so a point seen from $50$\,m away is
localized far more loosely than one seen from a few meters. A point-cloud
alignment that treats every point as equally certain then forces such a
loosely-triangulated far point, from one sequence, to line up precisely with a
tightly-triangulated near point of the same structure from another; without
propagating those per-point uncertainties into the cost, the solve trades a
well-localized constraint against a badly-localized one and settles at a
poorly-converged compromise. The method's own choices show that
without shared structure the estimate is not consistent enough, and with it the
``BA-free'' claim no longer holds.

\subsection{Consumer GPS is a noisy ruler, not an anchor}
\label{sec:why-gps}
GPS plays an outsized role: it is the \emph{only} metric signal, used to scale
each pairwise translation and to select neighbors, yet it enters only as a
\emph{relative} ruler on translations, never as a soft \emph{absolute} position
prior on the nodes. So nothing pins global scale or
arrests drift; each chunk's scale floats independently, exactly the freedom
\cref{sec:why-scale} cannot later remove.

Worse, the ruler itself is imprecise. These drones carry consumer-grade GNSS (GPS).
With an RTK fix the receiver is accurate to a few centimeters, but the captures
rarely hold an RTK lock, and without it horizontal position is off by
$0.5$--$1.0$\,m. That error does not stay contained: fed in as the scale of every
essential-matrix translation, it is large relative to the inter-frame baselines
it scales, injecting meter-scale noise into precisely the $\SEthree$ translations
the merge already cannot correct.
Altitude is worse still: on several of these drones the vertical channel comes
from a \emph{barometric} altimeter, so the relative-$Z$ component carries far more
uncertainty than the horizontal ones. Feeding a strongly \emph{anisotropic},
meter-scale, sometimes-biased position signal into an optimizer that treats it
as a metric constraint makes the pose graph ill-posed, though not uniformly. The
circular orbit sequences are largely safe: their wide, multi-directional
baselines triangulate structure, and vertical scale with it, from the imagery.
The damage concentrates on the \emph{nadir sequences}, the near-straight
vertical ascents flown looking straight down, where the geometry is degenerate:
straight-line motion gives no cross-baseline, and the nadir view aligns the
poorly-triangulated depth axis with the vertical, so ``the camera climbed'' and
``the scene grew'' are nearly indistinguishable. Vertical scale is then only weakly
constrained: the nadir edges resist it little, and the one term that penalizes a
wrong vertical scale, the barometric-$Z$ prior, is too noisy to push back. With
so little curvature there, \cref{eq:pgo} is nearly flat, so the estimate can
drift far at almost no cost. This
is a direct source of the large outliers and outright failure we observed
(\cref{tab:gauu}). A pose-only formulation has no shared structure to fall back on
when its metric signal is this noisy; bundle adjustment derives scale from the
imagery itself and needs GPS only as a weak gauge, so the same fix does it little
harm.

\subsection{Trust is hand-set, not propagated}
\label{sec:why-trust}
Compounding the noise, every constraint's influence is a hand-chosen information
matrix (rotation over translation, PnP $7\times$ essential, refined edges
$10\times$) rather than a value propagated from the data's uncertainty. Without a
data-driven model, a triangulation-noisy or badly-GPS-scaled match cannot be
down-weighted against a clean one, the mechanical reason the $3$D--$3$D
constraint was unusable and the reason the same fixed weights that worked on the
tuning scenes produced large outliers and an outright failure (HAV) on a capture
whose noise profile differed. Stability resting on manually balanced trust, over
a signal as heteroscedastic as consumer GPS, does not transfer. This is the most
fixable of the four causes and the one that most confounds our results: with no
learned-weight configuration run, we cannot separate error from the formulation
from error from weighting it badly.

\section{Lessons}
\label{sec:lessons}
The causes above translate into concrete guidance for anyone drawn to the same
shortcut, replacing structure-aware optimization with a cheaper pose-only
surrogate on structured captures.

\begin{enumerate}
\item \textbf{Two independent speedups, and the front end is the bigger one.}
Dropping the points for a structure-less, pose-only optimization is genuinely
faster than a full bundle adjustment: it removes the most expensive part of the
solve, and that saving is real and worth taking. The larger opportunity, though,
is the matching front end that dominates absolute cost. It is embarrassingly
parallel and shards cleanly across workers, and the levers that matter there are
mundane, the retrieval budget per frame, pruning pairs before verification, and
GPU-parallel extraction and matching. Both are worth accelerating, but the front
end's gains are far easier to capture, because they come from trivial parallelism
rather than from reformulating the estimator, and reformulating the estimator is
what costs accuracy.
\item \textbf{If you drop the points, add scale back explicitly.} BA's hidden
gift is that shared structure couples the scales of overlapping views. A
pose-only formulation must recover that coupling elsewhere, through $\SIMthree$
edges and a $\SIMthree$ merge estimated jointly over all sub-maps rather than
pairwise (\cref{sec:why-scale}). $\SEthree$ pose graphs across
independently-scaled sub-maps are a scale trap.
\item \textbf{Use GPS as an absolute prior, not merely a ruler, but model its
noise.} A soft absolute position constraint on every node pins global scale and
arrests drift. It helps only if weighted by its true, \emph{anisotropic}
uncertainty, looser in barometric $Z$ than in the horizontal plane and looser
still without an RTK lock, otherwise the meter-scale error re-enters as bias.
\item \textbf{Keep a little structure.} A pose graph with \emph{zero} shared 3D
points is on the wrong side of a cliff. Retaining even a small point set, a few
times the camera count rather than the hundreds-of-times BA carries, restores
multi-view consistency while keeping the problem small, the structured-but-lean
middle, not pure PGO, the cost argument should have pointed to.
\item \textbf{Propagate uncertainty; do not hand-tune it.} Data-driven
information matrices are what make structure-to-structure constraints usable.
Fixed trust ratios are a liability that grows with capture diversity.
\end{enumerate}

\section{Conclusion}
\label{sec:conclusion}
We asked whether, in the dense loop-closure regime splatting captures uniquely
provide, pose-graph optimization could displace bundle adjustment as the
estimation objective. For the system we built the answer is no, twice over: it
did not match \colmap{}'s accuracy, and it did not even dispense with BA to get
there, keeping a per-chunk pass whose triangulated points its own cross-chunk
constraints are built on.

The causes divide, and the division is the point. Some are ours to own: a
consumer-GPS anchor too noisy and too anisotropic to hold scale, greedy pairwise
merging, sequence-by-sequence ordering, and hand-set information matrices. Each
is a repairable engineering decision we did not ablate, and a careful
reimplementation should beat our numbers. Two causes are not repairable that
way. The $\SEthree$ state space cannot represent the scale differences its own
GPS-scaled sub-maps introduce, whatever weights the edges carry; and a graph of
two-view edges cannot reconstitute the many-view coupling that gives bundle
adjustment its consistency. These are properties of the formulation, not settings
within it, and they are what a future design has to answer for: $\SIMthree$
constraints estimated jointly rather than pairwise, absolute uncertainty-aware
GPS priors, and a thin layer of retained structure. How much of the measured gap
those limits account for would take the ablations we did not run; we hope the
account still saves the next team the same detour.

\paragraph{When the approach might be worth revisiting.} Several failures trace
to the noisy metric anchor, not the formulation. Under an RTK fix, position is
accurate to a few centimetres in all of $(x, y, z)$, enough to pin scale across
merges and stop inter-sequence constraints conflicting at the meter scale. A
lightweight distributed pose graph, with a \emph{few sparse retained points}
supplying the coupling pure $\SEthree$/$\SIMthree$ edges cannot, would then
deserve a fresh attempt: the negative result is a scoping one, pose-only,
GPS-scaled estimation is the wrong tool for \emph{consumer}-grade capture.

\FloatBarrier
\label{sec:bodyend}
\bibliographystyle{splncs04}
\bibliography{main}

\end{document}